\pdfoutput=1

\documentclass[11pt]{article}

\usepackage{emnlp2021}

\usepackage{times}
\usepackage{latexsym}
\usepackage{tikz}
\usepackage{ifthen}
\usepackage{adjustbox}
\usepackage{latexsym}
\usepackage{xcolor}
\usepackage{ifthen}
\usepackage{mathtools}

\usepackage{graphicx}
\usepackage{siunitx}

\usepackage{soul}
\usepackage{xcolor}
\usepackage{amsmath}
\usepackage{amsfonts}
\DeclareMathOperator*{\argmax}{arg\,max}

\usepackage{graphicx}
\usepackage{subcaption}

\usetikzlibrary{backgrounds,shadows,scopes,fit,calc,patterns}
\pgfmathsetseed{16}

\newcommand{\attrow}[7]{
    \pgfmathsetseed{#3}
    \begin{scope}
    \foreach \i in {1,2,...,#3}{%
        \pgfmathsetmacro{\left}{\i + 1}
        \ifdim#4 pt<\left pt
            \pgfmathsetmacro{\right}{\i - 1}
            \ifdim#5 pt>\right pt
                \pgfmathsetmacro{\currand}{0.8}
                \node [cell node, fill=#6, fill opacity=\currand] at (\i * \bs + #1, #2){};
            \else
                \node [cell node] at (\i * \bs + #1, #2){};
            \fi
        \else
            \node [cell node] at (\i * \bs + #1, #2){};
        \fi
    }
    \node [label node] at (#3 * \bs + #1 + \bs , #2){#7};
    \end{scope}
}

\usepackage{amsmath}
\usepackage{hyperref}

\newcommand{\bx}{\mathbf{x}}
\newcommand{\by}{\mathbf{y}}
\newcommand{\bz}{\mathbf{z}}

\newcommand{\rscomment}[1]{\textcolor{black}{#1}}
\newcommand{\sgcomment}[1]{\textcolor{black}{#1}}
\newcommand{\sgwording}[1]{\textcolor{black}{#1}}

\newcommand{\ie}{\textit{i.e.}}
\newcommand{\eg}{\textit{e.g.}}

\newcommand{\bea}{{\sc BoitoEA}}

\usepackage[T1]{fontenc}

\usepackage[utf8]{inputenc}

\usepackage{microtype}

\title{On the Difficulty of Segmenting Words with Attention}

\author{Anonymous submission}
\author{Ramon Sanabria~~~~Hao Tang~~~~Sharon Goldwater \\
School of Informatics\\
   The University of Edinburgh \\
   {\small\texttt{\{r.sanabria, hao.tang\}@ed.ac.uk, sgwater@inf.ed.ac.uk}} \\}

\begin{document}
\maketitle
\begin{abstract}
Word segmentation, the problem of finding word boundaries in speech, is of interest for a range of tasks.
Previous papers have suggested that for sequence-to-sequence models trained on tasks such as speech translation or speech recognition, \emph{attention} can be used to locate and segment the words.
We show, however, that even on monolingual data this approach is brittle.
In our experiments with different input types, data sizes, and segmentation algorithms, only models trained to predict phones from words succeed in the task. Models trained 
to predict words from either phones or speech (\sgwording{\ie, the opposite} direction needed to generalize to new data),
yield much worse results, 
\sgwording{suggesting that attention-based segmentation is only useful in limited scenarios.}\footnote{\rscomment{code available in the following link \href{https://github.com/ramonsanabria/insights_2021}{https://github.com/ramonsanabria/insights\_2021}}}

\end{abstract}

\section{Introduction}

Word segmentation is the task of finding word boundaries in speech.
The task has a wide range of applications,
including documenting under-resourced languages \cite{dunbar2017zero} \sgcomment{and}
bootstrapping speech recognizers \cite{juang1990segmental}.
\sgcomment{It is} often the first step to a variety of unsupervised
speech tasks \cite{chung2018speech2vec,chung2018unsupervised,baevski2021unsupervised}
\sgcomment{and to the NLP pipeline for languages with no whitespace between words.}

While unsupervised speech segmentation has been studied \cite{kamper2017segmental,rasanen2015unsupervised}, in many cases a parallel data source may be available, such as transcriptions, translations, or images. Previous researchers have suggested that it is possible to extract word segments from the attention map created by training an end-to-end sequence-to-sequence model on such parallel data \cite{palaskar2018acoustic,boito2017unwritten,boito2019empirical, boito2020investigating, godard2018unsupervised}. However, \citeauthor{palaskar2018acoustic}'s evaluation was non-standard\footnote{For their ASR model, they reported mean frame error relative to a forced alignment. Positive and negative errors cancel, so a small mean error does not imply correct boundaries.} and the others  (which we refer to collectively henceforth as \bea) used models with text translations on the \emph{input} side---decoding to phones or phone-like units---which means the trained models cannot be applied to segment novel (untranslated) sequences. In addition, the interpretation of attention as alignments in other areas of NLP has been questioned \cite{jain2019attention, wiegreffe2019attention}.

Prompted by the prior work, we set out to study
word segmentation with attention in sequence-to-sequence models, aiming to better understand when and how attention can be interpreted as alignments and in what settings it can be used to segment words.
Our main experiments follow \bea\ in performing and evaluating word segmentation on the same data used to train the sequence-to-sequence model. 
However, instead of training 
translation models as in \bea, we train models
to perform speech recognition:
a well-studied task with a simpler (monotonic) alignment structure. 
This setting is similar to forced alignment, where 
both the speech
and the transcription are given, and the goal is to discover
the hidden alignments. 
Aside from this potential use case, 
this setting is useful for analysis because it abstracts away from the need to generalize to a novel test set. 
If the attention is not able to provide acceptable word
boundaries in this setting, then the approach is unlikely to
succeed in other, more difficult, settings.

We perform experiments on both the low-resource Mboshi dataset from \bea\ and a much larger English dataset, MuST-C \cite{di2019must}. We study models trained in both directions with a variety of input-output types (\eg, phones, speech frames) and different postprocessing strategies to extract alignments from the attention weights. We find that although the particular configuration used by \bea\ works well, 
in most configurations the word segments extracted from the attention map are poor,
even when using a larger dataset or model size. In particular, we did not get good results from any of the configurations that can be applied to a novel test set or to speech frames (rather than phone-like units). We conclude that even in the simple monotonic case studied here, attention typically does not provide clear word-level alignments.

\section{Problem Setting}
\label{sec:prob}

We consider an $n$-sample dataset
$S = \{(\bx_1, \by_1, \bz_1), \dots, (\bx_n, \by_n, \bz_n)\}$,
each of which is a triplet $(\bx, \by, \bz) \in \mathcal{X} \times \mathcal{Y} \times \mathcal{Z}$.
As an example, $\mathcal{X}$ is the set of sequences of speech frames,
$\mathcal{Y}$ is the set of sequences of words,
and $\mathcal{Z}$ is the set of sequences of segments,
where a segment is a triplet $(s, t, w)$
that indicates the start time $s$, the end time $t$, and the word $w$.

The goal is to learn a function $f: \mathcal{X} \times \mathcal{Y} \to \mathcal{Z}$
given \emph{only} $S|_{xy} = \{(\bx_1, \by_1), \dots, (\bx_n, \by_n)\}$,
i.e., discovering the alignments
without observing them. Formally, we aim to find $f$ that minimizes
$\sum_{i=1}^n \ell(f(\bx_i, \by_i), \bz_i)$ using only $S|_{xy}$,
for some loss function $\ell$ that evaluates the quality of the segmentation.
Note that the evaluation of $f$ is on the set $S$.
We could evaluate $f$ on a test set, but, in general,
generalization is \emph{not} involved in this setting.\footnote{We do report generalization results in the Appendix, for completeness, though these do not change our main story.}

The setting is general, subsuming many tasks.
When $\mathcal{X}$ is the set of speech utterances
and $\mathcal{Y}$ is the empty set, this is the usual unsupervised word segmentation.
The set $\mathcal{Y}$ can be images or translations, grounding words from other modalities \cite{harwath2018jointly}. 
In this work, we focus on $\mathcal{Y}$ being transcriptions,
\ie, we have a forced alignment task.

\begin{figure}
  \centering

\begin{subfigure}{.4\textwidth}
  \centering
  \includegraphics[width=1.1\linewidth]{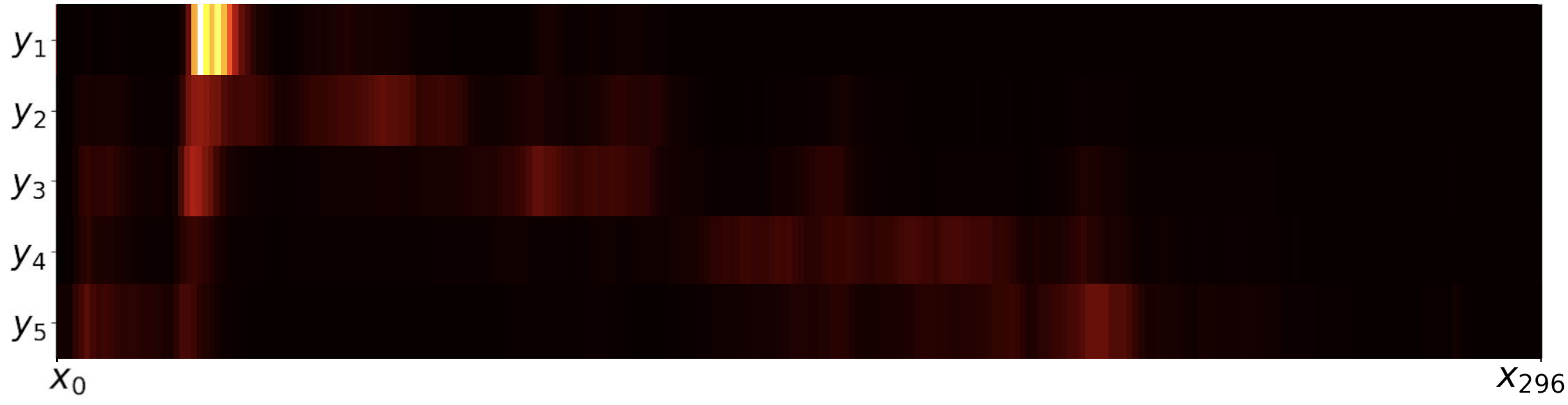}
   \caption{speech}
\label{fig:att_speech}
 \end{subfigure}
\begin{subfigure}{.4\textwidth}
  \centering
  \includegraphics[width=1.1\linewidth]{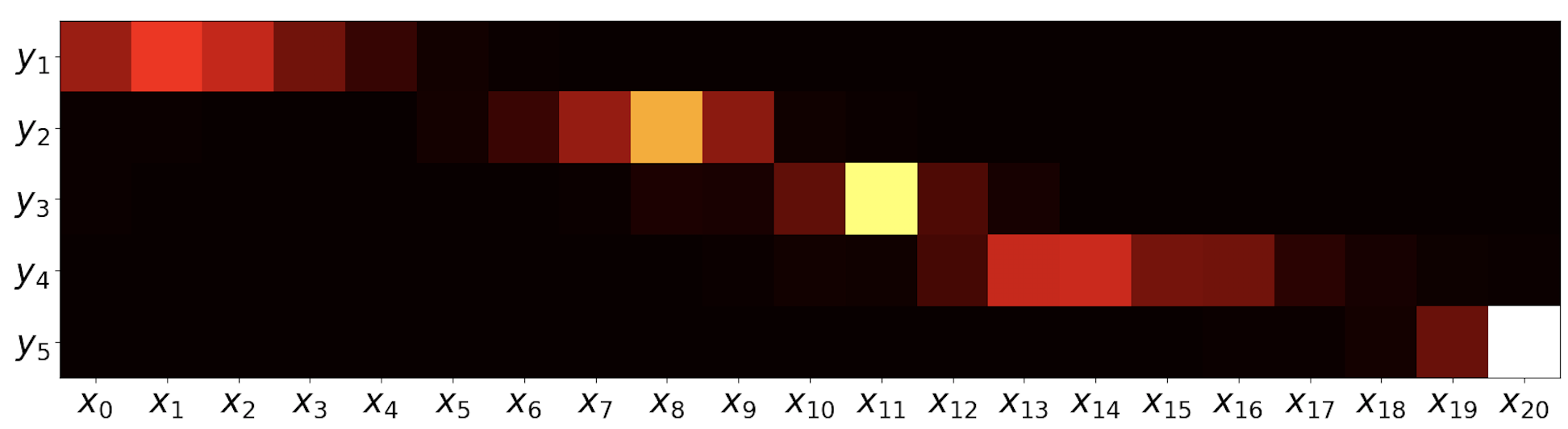}
   \caption{phones}
\label{fig:att_phn}
 \end{subfigure}

   \caption{\sgcomment{Example attention 
   maps for sequences using (a) speech or (b) phones as input. Note that for space reasons, the maps are shown transposed: each row shows the attention for a single output timestep.
   }}\label{fig:att}
\end{figure}

\section{Word Segmentation with Attention}
\label{sec:word_seg}

Our pipeline, due to \bea,
consists of two steps:
generating an attention map from a sequence-to-sequence model,
followed by postprocessing to 
convert the map into an alignment.

\subsection{Sequence-to-Sequence Models}

Below is a review of sequence-to-sequence models.
Readers should refer to, for example, \citet{luong2015effective} for a detailed exposition.

Given a speech utterance $\bx = x_1 x_2 \cdots x_T$ or simply $x_{1:T}$
and its transcription $\by = y_{1:K}$, a sequence-to-sequence model
learns a function of $\mathcal{X} \to \mathcal{Y}$.
An encoder $\text{Enc}$ and a decoder $\text{Dec}$
take $\bx$ and $\by$ as input
and produce their respective hidden vectors $h_{1:T}= \text{Enc}(x_{1:T})$ and $q_{2:K}= \text{Dec}(y_{1:K-1})$.

The attention map
\rscomment{
\begin{align}
    \alpha_{t, k} = \frac{\exp \Big( W_a \begin{bmatrix} h_t \\ q_k \end{bmatrix} \Big)}{\sum_{i=1}^T \exp \Big( W_a \begin{bmatrix} h_i \\ q_k \end{bmatrix} \Big)}
\end{align}
}

\rscomment{is computed with a weight matrix $W_a$. \sgcomment{Examples are illustrated in Figure \ref{fig:att}.} In the early stages of the project, we experimented with dot product attention and the results were similar.}
Finally, the probability of the label is computed as
\begin{align}
    p(y_k | y_{1:k-1}, x_{1:T}) = \text{softmax}\left(W \begin{bmatrix} c_k \\ q_k \end{bmatrix}\right)
\end{align}
where $c_k = \sum_{t=1}^T \alpha_{t, k} h_t$.
The model is trained to maximize the probability
\begin{align*}
    p(y_{1:K} | x_{1:T}) = p(y_1 | x_{1:T}) \prod_{k=2}^K p(y_k | y_{1:k-1}, x_{1:T}).
\end{align*}

We emphasize that, in our setting, $\bx$ and $\by$ are always given,
also known as teacher forcing \cite{lamb2016professor},
and we are interested in the attention map $\alpha$,
not how well the model maps $\bx$ to $\by$.
Note that the assignments to $\bx$ and $\by$ can be swapped since both are given.
For example, we can align word transcriptions to phonetic transcriptions, or vice versa.
However, the choice of directionality has two implications. First, 
for each output $y_k$, some parts of $\bx$ will have high attention weights, whereas for parts of $\bx$, there may be no $y_k$ with high weights. This asymmetry affects the choice of postprocessing method, as described below.
Second, typically only one direction will be feasible if we want to apply the trained model to new (unannotated) data. While our experiments here focus on segmenting the training set, we would ideally like to find a method that can also work on new data.

\subsection{Postprocessing}
\label{sec:postprocessing}

We explore three types of postprocessing to obtain the alignments from the attention map $\alpha$. 
The first of these was introduced by \citet{boito2017unwritten}; the others are novel.

\paragraph{Hard Assignment}
This approach aligns each output symbol $y_k$ to the input  that has the highest attention weight, i.e., to $x_{t_k}$, where $t_k = \argmax_t \alpha_{t, k}$.
Due to the attention map asymmetry noted above, this approach is only applied when the transcribed words are on the input side; otherwise some phones (or speech frames) may not be aligned to any word. 
This method hypothesizes a word boundary between two output symbols if they are aligned to different words; otherwise they are considered part of the same word.

\bea\ mainly use this method to train \emph{translation} models (French input words; Mboshi output phones), but \citet{boito2020investigating} also present monolingual results (Mboshi input words; Mboshi output phones), which we compare to below.

\paragraph{Thresholding}
When the attention weight is higher than a threshold
$\tau_{\text{onset}}$, then we hypothesize a start of a word segment.
When the attention weight is lower than a threshold
$\tau_{\text{offset}}$, then we hypothesize an end of a word segment. Thresholds are set by exhaustive search using F-score on the development set as the search metric. \rscomment{Thresholding can generate multiple segments for a given output, which is not desirable for our setting. However, in an automatic speech translation setup, such behavior can be helpful in some language pairs.}

\paragraph{Segmental Assignment}
Since we know that word segments are contiguous chunks of speech,
this constraint should be baked into the postprocessing.
In particular, we find a sequence $(s_1, t_1), \dots, (s_K, t_K)$
such that the sum of attention weights that each segment covers,
\ie, $\sum_{t=s_k}^{t_k} \alpha_{t, k}$, is maximized,
while respecting the connectedness constraint,
\ie, $s_{k+1} = t_k + 1$.
This can be achieved by finding the maximum weighted path
in a graph with edges as word
segments and weights of the edges as the attention weights
a segment covers. See a detailed description in Appendix \ref{sec:segmental} and \cite{tang2017end}.

\section{Experiments}

Most of our experiments are conducted on the Mboshi dataset. 
It contains 4616 short read-speech utterances for training (3 seconds/6 words on average; 4.5h in total), with a vocabulary of 6638 words, and 514 utterances for development. 
Mboshi is a Bantu Language with no orthography, and the speech is transcribed
at the word level using a phonetic orthography designed by linguists.
We regard the basic units in the transcriptions as phones.\footnote{\sgcomment{We use the term \emph{phone} rather than \emph{phoneme} both here and with reference to the MUST-C data set (below), to avoid making any commitments about the underlying cognitive/linguistic form, which the term \emph{phoneme} implies. For the Mboshi data especially, we are not sure if these commitments hold. However, like phonemic transcriptions, the transcriptions we work with assume a single pronunciation for each word type (effectively, dictionary lookup of pronunciations).}}
We do not use the French translations of the utterances.

\rscomment{For experiments with speech, we use Kaldi to extract speech features, with a 25 ms window shifted by 10 ms. Each acoustic frame consists of 40-dimensional log mel features and 3-dimensional pitch features.}
\sgcomment{The acoustic feature vectors are used directly, we do no clustering or acoustic unit discovery.}

We use a 1-layer bidirectional LSTM encoder and a 1-layer unidirectional LSTM decoder,\footnote{\citet{boito2020investigating} showed that LSTMs worked better than Transformer or CNN models with their framework.} with 0.5 dropout on the encoder and a 256-dimensional hidden layer.\footnote{The number of layers, dropout, and dimensions were tuned on the development set. The values we found best are the same ones \citet{boito2020investigating} reported, except their hidden layer size is 64. It is unclear why dropout helps in this setting, since without generalization, there is no concern of overfitting, but we did find a small benefit.}
Further hyperparameter details are in Appendix \ref{sec:reproduction}.  

\sgcomment{As noted above, our main questions do not require testing generalization, so except where otherwise noted (to compare to previous work), we evaluate all models on the training set. Results on the development set, which do not change the story, are reported in Appendix \ref{sec:devset}.} We report precision, recall, and F-score of the hypothesized word boundaries.
When transcribing phones to words, a word boundary must be hypothesized
at the exact place to be counted as correct.
In later experiments, particularly for speech,
we \rscomment{follow the The Zero Resource Speech Challenge \cite{dunbar2017zero} evaluation and} use a 30ms tolerance window, \ie, the hypothesized boundary
is counted as correct
if it falls withing 30 ms of the correct boundary. \rscomment{
Similar to \bea, we use force alignments extracted with a Kaldi \cite{kaldi} GMM-HMM model as ground truth word boundaries.}
We also report the amount of over-segmentation, defined as
$(N_h - N_{\text{ref}}) / N_{\text{ref}}$.
If the quantity is positive, the model hypothesizes too many boundaries; if the quantity is negative, the model hypothesizes too few boundaries.

\subsection{Predicting Phones From Words}
\label{sec:phn2wrd}

We begin by confirming the positive results from previous work, following \bea\ in training a model to predict phones given words. 

\begin{table}
\begin{center}
\caption{\label{tbl:reprod} Word boundary scores on Mboshi for models
predicting phones from words (w $\to$ p), as in 
\citet{boito2020investigating},
using Hard or Segmental assignment.
$^{\dagger}$Models trained on Train+Dev. ${}^*$Results from \citet{boito2020investigating}, averaging five attention maps.
    }

\begin{tabular}{lccccc}
    \hline
               &    & P    & R    & F    & OS   \\
    \hline
    w $\to$ p  & Hard & 92.3 & 83.2 & 87.5 & -9.8 \\
    w $\to$ p  & Seg & 93.5 & 93.5 & 93.5 & 0.0 \\
    \hline
     w $\to$ p$^{\dagger}$ & Hard & 95.5  & 85.7    & 90.4 & -10.2 \\
     w $\to$ p$^{\dagger*}$   & Hard & 92.9   & 92.1    & 92.5 & - \\
     \hline
  \end{tabular}
\end{center}
\vspace*{-\baselineskip}
\end{table}

Results for both Hard Assignment (Hard) and Segmental Assignment (Seg) are shown in Table~\ref{tbl:reprod}.
We transpose the attention matrix to run Seg so that a segment (word) can consist of multiple phones.
As expected, both methods work well, with the more principled Seg performing slightly better---though it has a slight advantage, since (due to teacher forcing) it always generates the right number of word segments (yielding OS = 0).

However, these methods can only be applied to the annotated data, which is a significant weakness.

\subsection{Words as Targets}

Next, we consider the more typical direction of decoding,
predicting words given phones.
We do not use hard assignment in this setting for the reasons described in Section \ref{sec:postprocessing}.

Results (Table~\ref{tbl:words}) show that, when using phones,
flipping the model direction from w~$\to$~p to p~$\to$~w makes the results much worse. 
The Thresholding method is especially bad, so we focus on Seg for the rest of the paper.
The deterioration by simply flipping the model
is unsatisfying, because the setting is simple enough that
the model should be able to achieve near-perfect results by acting like a lexicon, mapping canonical pronunciations to words.

This decoding direction allows us to build models
that take acoustic features as input and produce
words.

However, once we replace phone transcriptions
as input with acoustic features, the result
(denoted a~$\to$~w in Table~\ref{tbl:words})
is drastically worse---even underperforming one of the unsupervised baselines \rscomment{(note the change on attention structure between Figure \ref{fig:att_phn} and Figure \ref{fig:att_speech}) }.

To understand what causes the dramatic drop in performance,
we explore an intermediate input representation
where 
we replace each acoustic frame with its phone label. This input format, which we refer to as phone frames, has the same length as the acoustic input sequence and reflects the duration of each phone, while abstracting away from acoustic variability. 
Results of this experiment (f~$\to$~w in Table~\ref{tbl:words}) show
that most of the performance gap is recovered.
This suggests that the model
learns little about the phonetic variability in this experiment.
This prompts us to work on a larger dataset,
with the hope that the model is able to capture
the phonetic variability given more data.

\begin{table}
\begin{center}
\caption{\label{tbl:words} Word boundary scores on Mboshi for models with words as targets, using phones, phone frames, or acoustic feature frames as input (p $\to$ w, f $\to$ w, and a $\to$ w, respectively), with Thresholding or Segmental assignment.
    The first row is copied from Table~\ref{tbl:reprod}.
    Unsupervised baselines (acoustic input only) are also shown: R15 \cite{rasanen2015unsupervised}; K17 \cite{kamper2017segmental}. 
}
\begin{tabular}{lccccc}
    \hline
               &    & P    & R    & F    & OS   \\
    \hline
    w $\to$ p  & Seg & 93.5 & 93.5 & 93.5 & 0.0    \\
    \hline
    p $\to$ w  & Thr  & 19.2 & 20.4 & 19.8 & 6.3  \\
    p $\to$ w  & Seg & 58.0 & 58.0 & 58.0 & 0.0    \\
    \hline
    f $\to$ w  & Seg & 57.6 & 57.6 & 57.6 & 0.0    \\ 
    a $\to$ w  & Seg & 13.4 & 13.4 & 13.4 & 0.0    \\
    \hline
a & R15  &  21.5 &  19.8 &  20.6  & -8.1 \\
a  & K17  &  32.4 & 7.4 & 12.0 & -77.1  \\ \hline
\end{tabular}
\end{center}
\vspace*{-\baselineskip}
\end{table}

\subsection{Scaling to Larger Data}
\label{sec:large_data}

We investigate if by exposing a model to more speech data it would learn to normalize phonetic variance and close the gap between f $\to $ w and a~$\to$~w.
For this experiment we use English data (speech, lexicon phone sequences, and word transcriptions) from the MuST-C dataset \cite{di2019must}. MuST-C provides translations to other languages; we don't use these here but we do limit our data to the 145k English utterances (257h of speech) for which translations are available in all the languages\footnote{\rscomment{Find the list of files in Kaldi format in the following link  \href{https://homepages.inf.ed.ac.uk/s1945848/must_c_insights.zip}{https://homepages.inf.ed.ac.uk/s1945848/must\_c\_insights.zip}}.}.
Utterances have an average length of 6.5 seconds/18 words. \rscomment{We use the same speech feature extraction configuration as in the Mboshi experiments.}
Because we are using a larger dataset, we 
also try a deeper (5-layer) model for the speech input. 
Results are shown in Table \ref{tbl:mustc}. The results on p~$\to$~w are lower than for Mboshi, suggesting that the longer utterances in MuST-C make the task more challenging.
The speech in MuST-C is probably also harder than in Mboshi (TED talks vs.\ read speech);
nevertheless, the performance on a~$\to$~w is better on MuST-C than Mboshi, closer to the MuST-C f~$\to$~w results.
This suggests that adding more data does allow the model to learn more about acoustic variability. However, given the large size of this data set, all the results are underwhelming, and the results with speech still do not beat the unsupervised models.

\begin{table}
\begin{center}
\caption{\label{tbl:mustc} Word boundary scores on MuST-C using Segmental assignment with a variety of models. The first three models have 1 hidden layer, while Lg has 5. 
Unsupervised baselines are also shown: R15 \cite{rasanen2015unsupervised}; K17 \cite{kamper2017segmental} 
}

\begin{tabular}{lccccc}
    \hline
               &     & P    & R    & F    & OS   \\
    \hline
    p $\to$ w  & Seg & 44.0 & 44.0 & 44.0 & 0.0     \\
    f $\to$ w  & Seg &   42.6   &  42.6    &    42.6 &  0.0    \\ 
    a $\to$ w  & Seg  &  21.2   &   21.2   &    21.2  &      0.0 \\
    a $\to$ w (Lg) & Seg &   19.9   &  19.9   &  19.9    &    0.0  \\
    \hline
a&  R15  & 20.0 &  28.6 & 23.5  & 43.7 \\
a& K17  & 23.7 &  26.1 & 24.8  & 10.2 \\ \hline
\end{tabular}
\end{center}
\vspace*{-\baselineskip}
\end{table}

\section{Conclusion}

Previous researchers had suggested a connection between attention weights and word alignments in both speech recognition and speech translation. However, 
we have experimented with several attention-based segmentation methods and demonstrated that these only succeed in the scenario where words are used as the input to the model---a scenario with limited application. Performance drops considerably for models with phones as input, and is no better than unsupervised segmentation for models using speech as input, even when the amount of training data is increased by two orders of magnitude. Although in principle the transcriptions provide an additional source of information, using this to help segment words from speech will likely require a completely different approach.

\bibliography{anthology,custom}
\bibliographystyle{acl_natbib}

\newpage

\appendix

\section{Appendix}

\subsection{Segmental Assignment}
\label{sec:segmental}

Given an attention map, the goal of segmental assignment is to find a segmentation
that maximizes the amount of attention weights each word covers while
making sure that the word segments are connected.
To achieve this, we turn this into a problem of finding a maximum weighted path on a graph,
where the edges of the graph are segments,
the weights on the edges correspond to the amount attention weights covered,
and the graph encodes the connectedness constraints.

Suppose the attention map is of dimension $T \times K$.
Recall that $T$ is the number of input tokens (such as speech frames)
and $K$ is the number of output tokens (such as words).
We first create a vertex set $V = \{(t, k): \text{for $t = 0, \dots, T$ and $k = 0, \dots, K$}\}$, a grid marking every element in the attention map.
An edge is a pair of vertices $(t_1, k_1)$ and $(t_2, k_2)$
while satisfying $t_1 < t_2$ and $k_2 = k_1 + 1$.
That edge represents a segment of the $k_2$-th output token
that aligns to $t_1$ to $t_2$ on the input side.
This can be realized by defining the incoming edges
\begin{align}
\text{in}((t_2, k_2)) &= \Bigg\{\Big((t_1, k_2 - 1), (t_2, k_2)\Big): \notag \\
  & \quad \text{for $t_1 = 0, \dots, t_2$ if $k_2 > 1$} \Bigg\}.
\end{align}
We assign the sum of attention weights from $t_1$ to $t_2$,
\ie, $\sum_{t=t_1+1}^{t_2} \alpha_{t, k_2}$ to the edge $((t_1, k_2 - 1), (t_2, k_2))$.

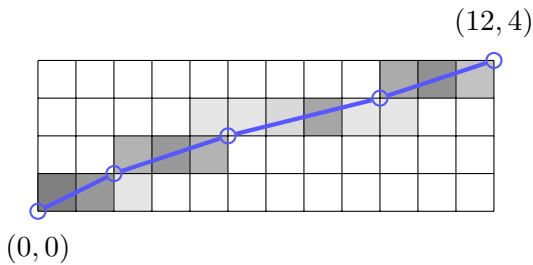
\begin{figure}
\begin{center}
\begin{tikzpicture}

\fill[black!50] (0, 0) rectangle (0.5, 0.5);
\fill[black!40] (0.5, 0) rectangle (1.0, 0.5);
\fill[black!10] (1.0, 0) rectangle (1.5, 0.5);

\fill[black!30] (1.0, 0.5) rectangle (1.5, 1.0);
\fill[black!40] (1.5, 0.5) rectangle (2.0, 1.0);
\fill[black!30] (2.0, 0.5) rectangle (2.5, 1.0);

\fill[black!10] (2.0, 1.0) rectangle (2.5, 1.5);
\fill[black!10] (2.5, 1.0) rectangle (3.0, 1.5);
\fill[black!15] (3.0, 1.0) rectangle (3.5, 1.5);
\fill[black!35] (3.5, 1.0) rectangle (4.0, 1.5);
\fill[black!10] (4.0, 1.0) rectangle (4.5, 1.5);
\fill[black!10] (4.5, 1.0) rectangle (5.0, 1.5);

\fill[black!33] (4.5, 1.5) rectangle (5.0, 2.0);
\fill[black!43] (5.0, 1.5) rectangle (5.5, 2.0);
\fill[black!24] (5.5, 1.5) rectangle (6.0, 2.0);

\draw[step=0.5, black, thin] (0, 0) grid (6, 2);

\node[draw, blue!66, thick, inner sep=2pt, circle] (a1) at (0, 0) {};
\node[draw, blue!66, thick, inner sep=2pt, circle] (a2) at (1, 0.5) {};
\node[draw, blue!66, thick, inner sep=2pt, circle] (a3) at (2.5, 1.0) {};
\node[draw, blue!66, thick, inner sep=2pt, circle] (a4) at (4.5, 1.5) {};
\node[draw, blue!66, thick, inner sep=2pt, circle] (a5) at (6, 2) {};

\draw[blue!66, ultra thick] (a1) -- (a2);
\draw[blue!66, ultra thick] (a2) -- (a3);
\draw[blue!66, ultra thick] (a3) -- (a4);
\draw[blue!66, ultra thick] (a4) -- (a5);

\node at (0, -0.5) {$(0, 0)$};
\node at (6, 2.5) {$(12, 4)$};

\end{tikzpicture}
\end{center}
\caption{\label{fig:seg} An example segmental assignment with 12 input tokens and 4 output tokens.
  Each box represents an element in the attention map.
  The darker the shade, the higher the attention weight.
  Edges of the maximum scoring path and the corresponding vertices are shown in blue.
  The shade being crossed by an edge is the amount of attention weights covered by the edge.
  The goal is of segmental assignment is to find the maximum weighted path from the bottom-left corner to the top-right corner. }
\end{figure}

Once the graph is constructed, we find the maximum weighted path
that starts at $(0, 0)$ and ends at $(T, K)$.
An example is shown in Figure~\ref{fig:seg}.
Note that due to the imposed constraints and
in turn due to how the graph is constructed,
segmental assignment only considers monotonic alignments.

\subsection{Hyperparameters and Implementation}
\label{sec:reproduction}

We use the sequence-to-sequence implementation of nmtpytorch \cite{nmtpytorch}\footnote{\href{https://github.com/lium-lst/nmtpytorch}{https://github.com/lium-lst/nmtpytorch}}. The model comprises one-layer encoder and one-layer decoder with 0.5 dropout, except in the experiment of Section \ref{sec:large_data} where we use a five-layer encoder in the Large (Lg) model. We set a size of 256 to all hidden dimensions (\ie, source and target embedding, encoder, and decoder). We use the Adam optimizer, and the scheduler applies a decay factor of 0.5 after two consecutive epochs where loss does not decrease. All models are trained until cross-entropy loss on training reaches 0. The implementation of each model has around 3M and 19M of learnable parameters for the 1 and 5 layers encoder model, respectively. They are trained with one Nvidia GEFORCE GTX 1080 Ti. \rscomment{To reduce computation on Segmental Assignment, we set the maximum duration of a word to 4 seconds (400 frames for speech or phone frames representation) for Mboshi and 10 seconds for MuST-C.} We set them by analyzing their performance on the development set.

Regarding the unsupervised speech baseline models, we use the unigram public implementation of \citet{kamper2017segmental}\footnote{\href{https://github.com/kamperh/bucktsong_segmentalist}{link to Kamper et al., 2017 Github implementation}} with a minimum word segment duration of 250 ms. Because its performance is linked to the syllable segmentation method, we select the best configuration by fine-tuning \citet{rasanen2015unsupervised}'s\footnote{\href{https://github.com/kamperh/recipe_zs2017_track2/tree/master/syllables/thetaOscillator}{link to Rasanen et al., 2015 Github implementation}} hyperparameter values on the development set. 

\subsection{Results without Transcriptions}
\label{sec:devset}
\begin{table}\caption{\label{tbl:unseen_results} Results from 1 layer, and 5 layers (Lg) models for word segmentation of the development (unseen) set of the Mboshi and MuST-C dataset.}  
\centering
\begin{tabular}{lcccccc}
DS & Model & P & R  & F & OS (\%) \\ \hline
 Mb & p $\to$   w &  51.9 & 54.2 & 53.0 & 4.4   \\
 Mb & f $\to$   w &  48.5 & 47.3 & 47.9 & -2.3 \\
  Mb & a  $\to$  w &   16.2 & 14.7 & 15.4 & -8.9  \\ 
 Mb & a  $\to$  w (Lg) &   14.1 & 14.9 & 14.5 & 6.0  \\ 
 \hline
  MC & p $\to$   w &  43.8  & 44.0 & 43.9 &  0.5  \\
 MC & f $\to$   w &  30.1 & 30.1 & 30.1 &  0.2 \\
 MC & a  $\to$  w (Lg) &   17.9 & 20.7 & 19.2 & 15.8 \\ \hline
\end{tabular}
\end{table}

Finally, we consider a more traditional scenario where the model is exposed to unseen data. For this setting our model does not have access to transcriptions and therefore we do not use Teacher-Forcing. We evaluate the development set from Mboshi (used in Section \ref{sec:phn2wrd}), which has 1147 token types (where only 710 are observed during training). In terms of absolute tokens, it has 2993, and only 516 out-of-vocabulary. We experiment with the MuST-C by using 1162 utterances from MuST-C's superset unseen during training. In this case, the set has 3273 token types. 

Surprisingly, for p and f $\to$ w, attention still produces meaningful segments although the model has not seen or early stopped with a development set. In that case, we observe a degradation in performance but not dramatic. The small number of unobserved absolute tokens do not have a critical effect on the segmentation performance of the model. 

Finally, the small difference in performance between the train and development sets on a $\to$ w shows an already present weak segmentation signal not correlated with word in speech models.

\end{document}